\pdfoutput=1
\documentclass[11pt]{article}

\usepackage[]{acl}

\usepackage{times}
\usepackage{latexsym}

\usepackage[T1]{fontenc}
\usepackage[utf8]{inputenc}

\usepackage{microtype}
\usepackage{tikz,pgfplots}

\usepackage{graphicx}

\usepackage[linesnumbered,ruled,vlined]{algorithm2e}
\SetKwRepeat{Do}{do}{while}%
\SetKwProg{Fn}{Function}{}{}
\SetKwComment{Comment}{$\triangleright$\ }{}
\SetArgSty{textnormal}
\usepackage{xcolor}

\SetCommentSty{mycommfont}

\usepackage{multirow, multicol}
\usepackage{booktabs}       % professional-quality tables
\usepackage{tikz}
\usepackage{xcolor}
\newcommand*\circled[1]{\tikz[baseline=(char.base)]{
            \node[shape=circle,fill,inner sep=2pt] (char) {\textcolor{white}{#1}};}}
\usepackage{amsmath}

\title{Bottom-Up Constituency Parsing and Nested Named Entity Recognition with Pointer Networks}

\author{Songlin Yang, Kewei Tu\thanks{\; Corresponding Author}\\
  School of Information Science and Technology, ShanghaiTech University \\
    Shanghai Engineering Research Center of Intelligent Vision and Imaging\\ 
    {\tt \{yangsl,tukw\}@shanghaitech.edu.cn}\\
 }

\begin{document}
\maketitle
\begin{abstract}
Constituency parsing and nested named entity recognition (NER) are similar tasks since they both aim to predict a collection of nested and non-crossing spans. 
In this work, we cast nested NER to constituency parsing and propose a novel pointing mechanism for bottom-up parsing to tackle both tasks. The key idea is based on the observation that if we traverse a constituency tree in post-order, i.e., visiting a parent after its children, then two consecutively visited spans would share a boundary. Our model tracks the shared boundaries and predicts the next boundary at each step by leveraging a pointer network. As a result, it needs only linear steps to parse and thus is efficient. It also maintains a parsing configuration for structural consistency, i.e., always outputting valid trees. 
Experimentally, our model achieves the state-of-the-art performance on PTB among all BERT-based models (96.01 F1 score) and competitive performance on CTB7 in constituency parsing; and it also achieves strong performance on three benchmark datasets of nested NER: ACE2004, ACE2005, and GENIA \footnote{Our code is publicly available at \url{https://github.com/sustcsonglin/pointer-net-for-nested}}.

\end{abstract}

\section{Introduction}

Constituency parsing is an important task in natural language processing, having many applications in downstream tasks, such as semantic role labeling \cite{fei-etal-2021-better}, opinion mining \cite{xia-etal-2021-unified}, among others. Named entity recognition (NER) is a fundamental task in information extraction and nested NER has been receiving increasing attention due to its broader applications \cite{DBLP:conf/semco/Byrne07}.

\begin{figure}[tb!]
    \centering
    \includegraphics[width=1\linewidth]{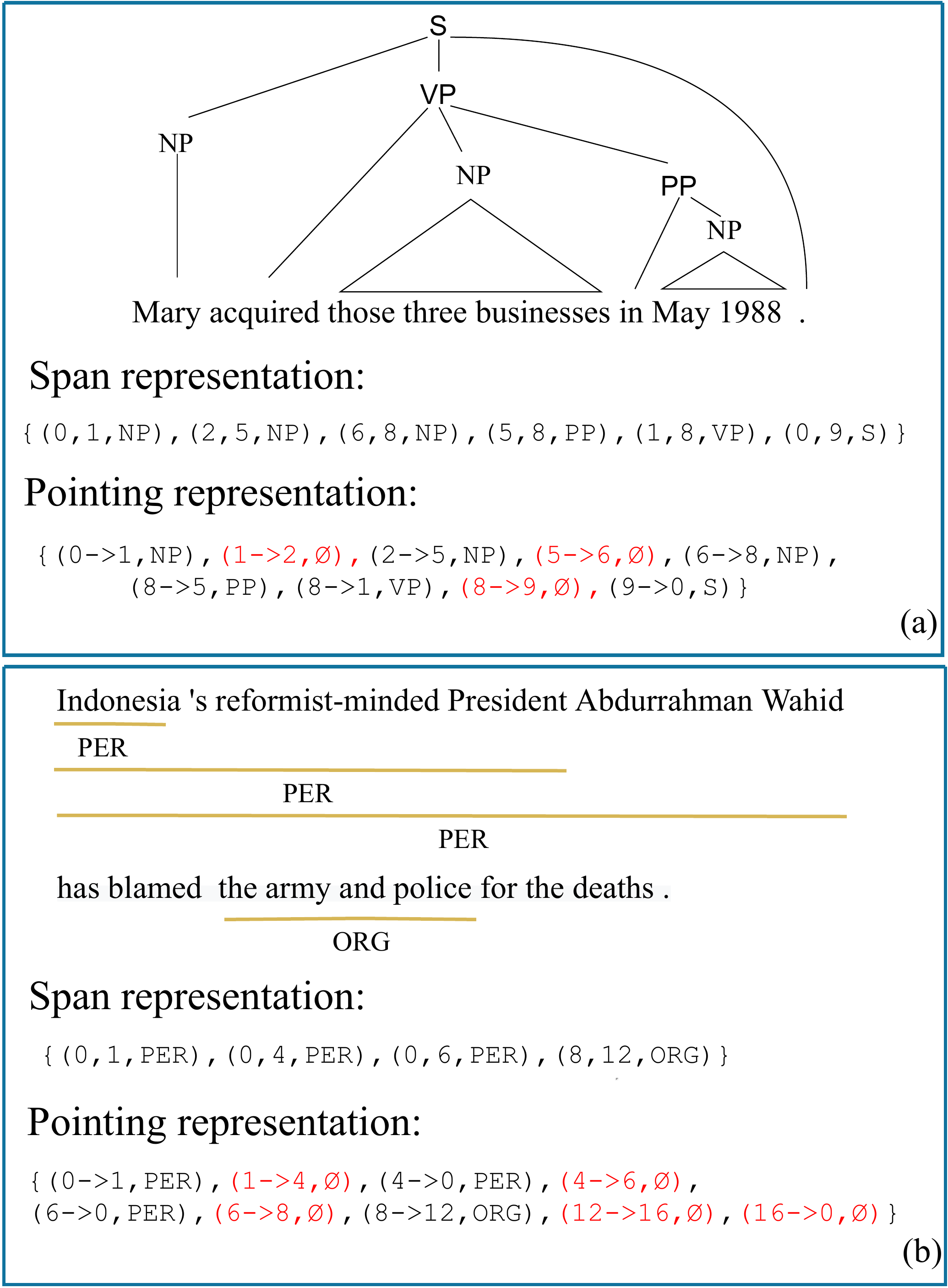}
    \caption{(a) an example non-binary constituency tree. (b) an example sentence with nested named entities. We show the span and pointing representations.}
    \label{fig:example}
\end{figure}

Constituency parsing and nested NER are similar tasks since they both aim to predict a collection of nested and non-crossing spans (i.e., if two spans overlap, one must be a subspan of the other). Fig.\ref{fig:example} shows example span representations of both tasks. The difference between the two tasks is that the collection of spans form a connected tree in constituency parsing, whereas they form several tree fragments in nested NER. However, we can add a node that spans the whole sentence to connect all tree fragments in nested NER to form a tree. Because of the similarity, there are some previous studies adapting methods from the constituency parsing literature to tackle nested NER \cite{finkel-manning-2009-nested, wang-etal-2018-neural-transition, TreeCRFNER}. In this work, we focus on constituency parsing, but our proposed method tackles nested NER as well.

The two main paradigms in constituency parsing are span-based and transition-based methods.  Span-based methods \cite[][\it{inter alia}]{stern-etal-2017-minimal, kitaev-klein-2018-constituency, TreeCRF, xin-etal-2021-n} decompose the score of a constituency tree into the scores of constituent spans and use chart-based algorithms for inference. Built upon powerful neural encoders, they have obtained state-of-the-art results.
However, they suffer from the high inference time complexity of exact algorithms or error propagation of top-down approximate algorithms. In contrast, transition-based methods  \cite[][\it{inter alia}]{dyer-etal-2016-recurrent,cross-huang-2016-span, liu-zhang-2017-order} conduct a series of local actions (e.g., shift and reduce) to build the final parse in linear steps, so they enjoy lower parsing time complexities. However, they suffer from the error propagation and exposure bias problems.

\begin{figure}[tb!]
    \centering
    \includegraphics[width=1\linewidth]{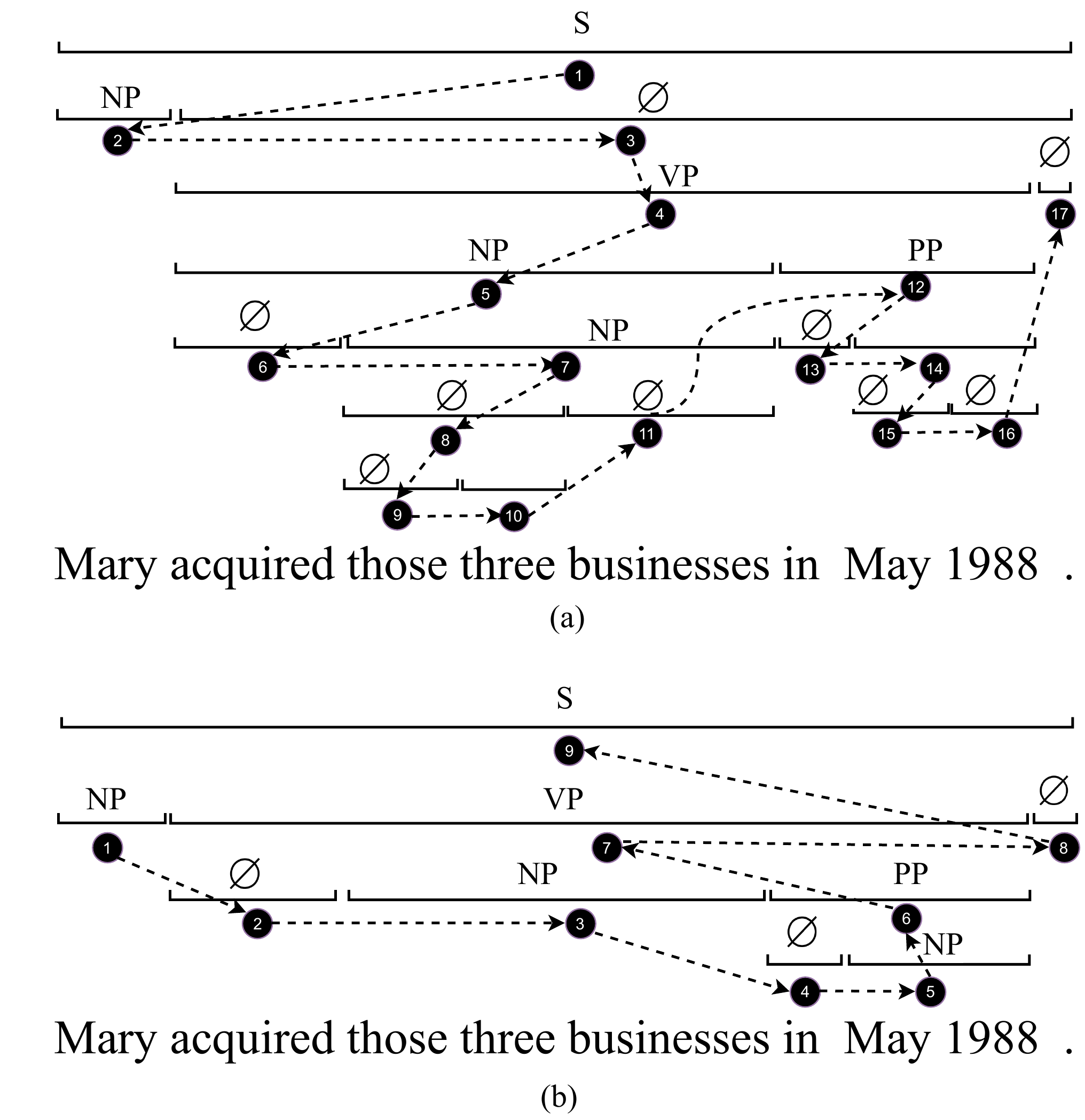}
    \caption{Illustration of pre-order and post-order traversal over the constituency tree shown in Figure \ref{fig:example}(a). (a): pre-order traversal. (b): post-order traversal. 
    We mark the generation order in the circles below spans and link two consecutively visited constituents by arrows. Note that in (a), binarization is assumed.}
    % Red segments denote $\emptyset$ spans that have no constituent labels.}
        \label{fig:post}
\end{figure}

Recently, \citet{nguyen-etal-2021-conditional} propose a sequence-to-sequence (seq2seq) model with pointer networks \cite{PointerNet}. They cast constituency parsing to a top-down splitting problem. First, they use neural encoders to obtain span representations, similar to span-based methods. Then
they feed input parent span representations into the neural decoder recursively following the order shown in Fig. \ref{fig:post}(a)\footnote{Slightly different from the figure, they do not feed spans of length 1 into the decoder for obvious reasons.}---which amounts to pre-order traversal---to output a series of splitting points (i.e., boundaries) via pointer networks, so that each parent span is split into two child spans. Notably, \citet{nguyen-etal-2020-efficient} propose a similar top-down pointing mechanism, but they 
design a chart-based parsing algorithm instead of adopting seq2seq modeling, and has been shown underperforming \citet{nguyen-etal-2021-conditional}. Thanks to seq2seq modeling, \citet{nguyen-etal-2021-conditional}'s model achieves a competitive parsing performance with a lower parsing complexity compared with span-based methods.
 
 However, their model has two main limitations. First, when generating each constituent, its subtree features cannot be exploited since its subspans have not been realized yet \cite{liu-zhang-2017-order}. Thus it is difficult for the model to predict the splitting point of a long span due to a lack of its subtree information, which exacerbates the error propagation problem and undermines the parsing performance. Second, since each parent span can only be split into two, their parsing algorithm can only ouput binary trees, thus needing binarization.
 
% However, when generating parent constituents, their subtree features cannot be exploited since their sub-constituents have not been realized, so intuitively it would be difficult to predict the splitting point of a long span and thus the error propagation problem would be exacerbated. In addition, their conditional splitting mechanism can only output binary trees, thus binarization is needed in preprocessing. 

 In this work, we devise a novel pointing mechanism for \textit{bottom-up} parsing using (almost) the same seq2seq backbone as \citet{nguyen-etal-2021-conditional}.
 Our model is able to overcome the two aforementioned limitations of  \citet{nguyen-etal-2021-conditional}. 
The main idea is based on the observation that if we traverse a constituency tree in post-order (i.e., visiting a parent after its children), two consecutively visited constituent spans would share a boundary. Fig. \ref{fig:post}(b) shows an example: the right boundary of \circled{1} is also the left boundary of \circled{2} and the right boundary of \circled{5} is also the right boundary of \circled{6}. Based on this observation, we propose to use a cursor to track the shared boundary boundaries and at each step, leverage a pointer network to predict the next boundary for generating the next constituent span and update the cursor to the right boundary of the new span. Our model generates one span at each step, thus needing only linear steps to parse a sentence, which is efficient. Besides, our model can leverage rich subtree features encoded in the neural decoder to generate parent constituent spans, which is especially helpful in predicting long spans. Finally, our model can output n-ary trees, enabling direct modeling of the original non-binary parse tree structures in treebanks and eliminating the need for binarization. 

% Compared with \citet{nguyen-etal-2021-conditional}, our method can leverage rich subtree features (encoded in the neural decoder) to generate parent constituent spans, which is especially helpful in predicting long spans; and can output n-ary trees, enabling direct modeling of the original non-binary tree structures within treebanks and eliminating the need of binarization. 

% Compared with span-based methods, our method has a lower parsing complexity and achieves competitive or even better performances. Besides, our model can leverage rich subtree information encoded in the neural decoder to disambiguate parsing, whereas span-based method can only utilize relatively local span representations for scoring.

% In addition, most previous constituency parsers require binarization in the preprocessing stage. In contrast, our method performs \textit{n-ary tree parsing} since it allows merging several words or subspans in a single step. The benefit of \textit{n-ary tree parsing} is that it is capable of modeling the original non-binary parse tree structures and does not need language-specific rules for binarization \cite{xin-etal-2021-n}.

We conduct experiments on the benchmarking PTB and CTB for constituency parsing. On PTB, we achieve the state-of-the-art performance (96.01 F1 score) among all BERT-based models. On CTB, we achieve competitive performance. We also apply our method to nested NER and conduct experiments on three benchmark datasets: ACE2004, ACE2005, and GENIA.  Our method achieves comparable performance to many tailored methods of nested NER, beating previous parsing-based methods. Our contributions can be summarized as the following:

\begin{itemize}
    \item We propose a novel pointing mechanism for bottom-up n-ary tree parsing in linear steps. 
    \item Our model achieves the state-of-the-art result on PTB in constituency parsing. We further show its application in nested NER where it achieves competitive results.
\end{itemize}

\section{Methods}

\subsection{Preprocessing}
It is known that constituency parsing can be regarded as a top-down splitting problem where parent spans are recursively split into pairs of subspans
\cite{stern-etal-2017-minimal, shen-etal-2018-straight, nguyen-etal-2020-efficient, nguyen-etal-2021-conditional}. However, this formulation can output binary trees. We make an extension to cast constituency parsing as top-down segmentation, i.e., parent spans are segmented into $\ge$ 2 subspans recursively, for the sake of outputting n-ary trees. To this end, we add some $\emptyset$ spans (we do not allow two adjacent $\emptyset$ spans to eliminate ambiguities) so that each span is either a bottommost span or can be segmented by its subspans. For instance, in Fig \ref{fig:post}, \circled{3} is a bottom-most span, and \circled{7} can be segmented by \circled{2}, \circled{3} and \circled{6}.
We always include the whole-sentence span in order to cast other tasks, e.g., nested NER, to constituency parsing.
We also collapse unary chains to atomic labels in constituency parsing, e.g., $\texttt{S->VP} \rightarrow \texttt{S+VP}$.

\subsection{Parsing configuration}

\begin{table*}
\begin{center}
\begin{tabular}{lrll}
    \toprule
    Initial configuration   & \multicolumn{3}{l}{$(c, A, p, s) = (0, \{1, 2, \dots, n \}, \texttt{null}, \emptyset)$} \\
    Goal       &  $(0, n) \in S$ \\
    \midrule
    Pointing action      & Input & Output & Precondition\\
    \midrule
    \textsc{Left-point}-$a$          & $(c, A, p , S)$
                            & $\Rightarrow (c, A \setminus \{a, \dots, c-1\}, a, S \cup \{(a, c)\})$ 
                            & $0 \le a < c$\\
    \textsc{Right-point}-$a$    & $(c, A, p, S)$
                            & $\Rightarrow (a, A \cup \{p\} \setminus \{c, \dots, a-1\} , c, S \cup \{(c, a)\} )$ 
                            & $ c < a \le n$,\\
    \bottomrule
\end{tabular}

\end{center}
\caption{Description of the parsing configuration.}
\label{tab:pointing-system}
\end{table*}
A problem of seq2seq constituency parsers is how to maintain structural consistency, i.e., outputting valid trees. To solve this problem, our pointing system maintains a \textit{parsing configuration}, which is a quadruple $(c, A, p, S)$ where:
\begin{itemize}
    \item $c$: index of the cursor.    \item $A$: set of indices of all candidate boundaries. 
    \item $p$: the left boundary of the lastly created span, which is needed to maintain $A$.
    \item $S$: set of generated spans. 
\end{itemize}

\begin{figure*}[tb!]
    \centering
    \includegraphics[width=\linewidth]{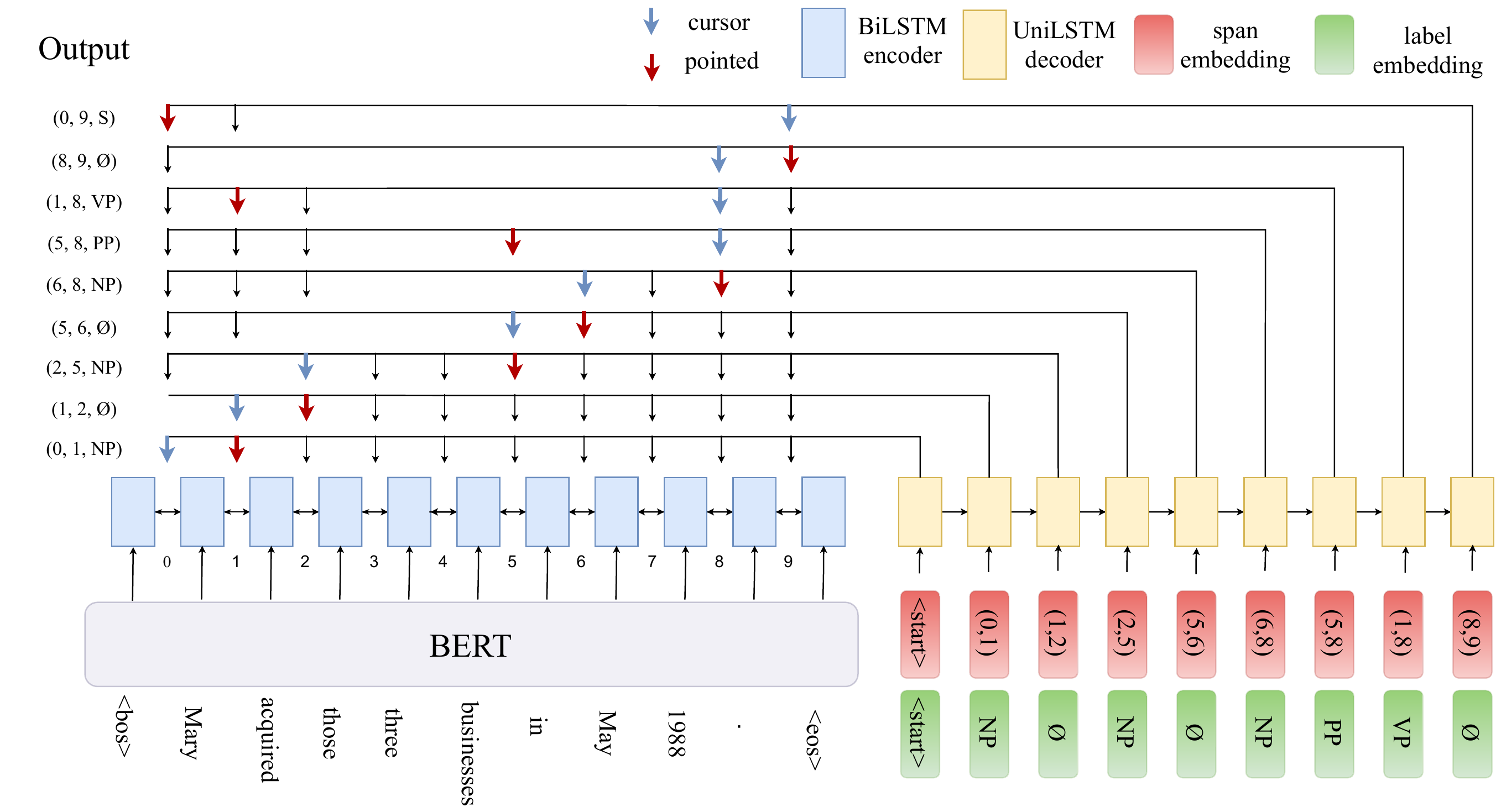}
    \caption{Demonstration of the generation process and the neural architecture. Black arrows point to candidate boundaries that are not selected in each step. }
    \label{fig:model}
\end{figure*}

We can see from Fig. \ref{fig:model} that in the beginning, the cursor $c$ lies at 0. At each step, $c$ points to another boundary $a$ from $A$ to form a span $(\min(c, a), \max(c, a))$. There are two cases:

\begin{itemize}
    \item $c<a$: a new bottom-most span is generated.
        \item $a<c$: several consecutive spans are merged into a larger span.  It is worthy to note that we can merge $>=2$ spans in a single step, which allows our model to perform n-ary tree parsing.
\end{itemize}

In the first case, the new bottom-most span can combine with the very previous span to form a larger span whose left boundary is $p$, so we push  $p$ back to $A$ (except for the case that $p=\texttt{null}$). In the later case, the very previous span is a subspan of the new span and thus $p$ cannot be pushed back. In both cases, all indices $\min(c,a) \le i < \max(c,a)$ are removed from $A$ due to the post-order generation restriction;  $p$ is updated to $\min(c,a)$  and $c$ is updated to $\max(c, a)$. The process stops when the whole-sentence span is generated. Table \ref{tab:pointing-system} formalises this process.

\paragraph{Oracle.} The oracle pointing representations shown in Fig.\ref{fig:example} can be generated by running a post-order traversal of the tree (e.g., Fig.\ref{fig:post}) and for each traversed span, pointing the cursor from its boundary shared with the previous span to its other boundary. If we do not allow two consecutive $\emptyset$ spans, the oracle is unique under our pointing system (we give a proof in Appendix A.1 by contradiction).

% \begin{algorithm}[tb!]
%   \KwData{spans $s=[(l_i,r_i,y_i)]_{i=1,...,m}$}
%   \KwResult{Pointing actions}
% % \Comment*[r]{Add auxiliary spans to $s$}
% actions $\leftarrow []$ \;
%  \tcc{Python-styled sorting}
% sorted(s,key=lambda x:(x[1],-x[0]))\label{sort2}\;
%  cursor $\leftarrow$ 0\; \label{focus}
% \ForEach{x $\in$ s}{
% \eIf{x[0] = cursor}{
%  push(actions, (x[0] $\rightarrow$ x[1], x[2]))\;
%  cursor $\leftarrow$ x[1]  
% %  \Comment*[r]{Point to the right and move the cursor.} 
% }
% {
%  push(actions, (x[1] $\rightarrow$ x[0], x[2])) 
% %  .\Comment*[r]{Point to the left.}
% }
% }\label{focusend}
% \Return actions
% \caption{Generating oracle actions.}
% \label{alg:oracle}
% \end{algorithm}

% \begin{equation}
% \texttt{sorted(s,key=lambda x:(x[1],-x[0]))}
% \end{equation}

\subsection{Model}
Given a sentence $w=w_1, ..., x_n$, we add \texttt{<bos>} (beginning of sentence) as $w_0$ and \texttt{<eos>} (end of sentence) as $w_{n+1}$. The oracle is $\{ q_i \rightarrow p_i, y_i\}_{i=1,...,m}$, where $y_i$ is the span label and
we use $l_i = \min (q_i, p_i)$ and $r_i = \max (q_i, p_i)$ to denote the left and right boundary of the $i$-th span, respectively. 

\paragraph{Encoder.} We feed the sentence into BERT \cite{devlin-etal-2019-bert} and for each word $w_i$, we use the last subtoken emebedding of the last layer as its dense representations $x_i$. Then we feed $x_0, \dots, x_{n+1}$ into a three-layer bidirectional LSTM \cite{hochreiter1997long} (BiLSTM) to obtain $c_0, \dots, c_{n+1}$, where $c_i = [f_i; g_i]$ and $f_i$ and $g_i$ are the forward and backward hidden states of the last BiLSTM layer at position $i$ respectively. 

\paragraph{Boundary and span representation.}
We use fencepost representation \cite{cross-huang-2016-span, stern-etal-2017-minimal} to encode the $i$-th boundary lying between $x_i$ and $x_{i+1}$:
\[
b_i = [f_i; g_{i+1}]
\]
then we represent span $(i, j)$ as:
\[
h_{i, j} = \text{MLP}_{\text{span}}(b_j - b_i)
\]

\paragraph{Decoder.} We use a unidirectional one-layer LSTM network as the decoder:
\begin{equation}
   d_t = \text{LSTM}(d_{t-1}, h_{l_{t-1},r_{t-1}} ; E_{y_{t-1}}), t\ge 2
   \label{eq:1}
\end{equation}
where $d_t$ is the hidden state of the LSTM decoder at time step $t$, $E$ is the label embedding matrix, $;$ is the concatenation operation.
For the first step, we feed a randomly initialized trainable vector $d_0$ and a special \texttt{<START>} embedding into the decoder to obtain $d_1$.

\paragraph{Pointing score.}  We use a deep biaffine function \cite{Biaffine} to estimate the pointing score $s^{t}_{i}$ of selecting the $i$-th boundary at time step $t$:
\begin{align*}
	d^{\prime}_t &= \text{MLP}_{\text{cursor}}(d_t) \\ 
	b^{\prime}_i &= \text{MLP}_{\text{point}}(b_i) \\
	s_{i}^{t} &=\left[b_{i}^{\prime} ; 1\right]^{\top}W_{\text{point}} d_{t}^{\prime}
\end{align*}
where $\text{MLP}_{\text{cursor}}$ and $\text{MLP}_{\text{point}}$ are multi-layer perceptrons (MLPs) that project decoder states and boundary representations into $k$-dimensional spaces, respectively; $W_{\text{point}} \in \mathcal{R}^{(k+1) \times (k)}$. 
  \paragraph{Label score.}
For a newly predicted span, we feed the concatenation of the span representation and the decoder state into another MLP to calculate the label score $e^{t}$:
\begin{align*}
% 	d^{\prime}_t &= \text{MLP}_{\text{focus}}(d^t) \\ 
	H &= \text{MLP}_{\text{label}}([ d^{t} ; b_{r_{t}} - b_{l_{t}} ]) \\
	e^{t} &= HE^{T}
\end{align*}
Note that we reuse the label embedding matrix from Eq. \ref{eq:1} to facilitate parameter sharing.

\paragraph{Training objective.} The training loss is decomposed into the pointing loss and the labeling loss:
\begin{align*}
    L &= L_{\text{pointing}} + L_{\text{labeling}} \\
    L_{\text{pointing}} &= - \sum_{t=1}^{m} \log \frac{\exp\{s_{p_{t}}^{t}\}}{\sum_{j=0}^{n}\exp\{s_{j}^{t}\}} \\
    L_{\text{labeling}} &= - \sum_{t=1}^{m} \log \frac{\exp\{e_{y_{t}}^{t}\}}{\sum_{j=1}^{|L|}\exp\{e_{j}^{t}\}}
\end{align*}
where $|L|$ is the number of labels.  Note that in the pointing loss we normalize over all boundaries instead of only accessible boundaries, because we find it performs better in our preliminary experiments. 

 \paragraph{Parsing.} Our model follows the description in the previous subsection for parsing. For each time step $t$, it selects the highest-scoring accessible boundary to generate the span, then selects the highest-scoring label of the generated span, and updates the parsing configuration (Table \ref{tab:pointing-system}).

% efficient beam search?

\section{Experiment setup}
\subsection{Data setup}
\paragraph{Constituency parsing.} We conduct experiments on Penn Treebank (PTB) 3.0 \cite{marcus-etal-1993-building} and Chinese Treebank (CTB) \cite{CTB}.  Many previous researchers report that the results on CTB5.1 
are unstable and of high variance \cite{TreeCRF,IncrementalParser}. So we follow the suggestion of \citet{TreeCRF} to conduct experiments on CTB7 instead of CTB5.1 for more robust evaluation as CTB7 has more test sentences and has a higher annotation quality. We use the standard data splits for both PTB and CTB.

\paragraph{Nested NER.} We conduct experiments on three benchmark datasets: ACE2004 \cite{doddington-etal-2004-automatic}, ACE2005 \cite{ACE05}, and GENIA \cite{GENIA}. We use the same data preprocessing as \citet{shibuya-hovy-2020-nested} \footnote{\url{https://github.com/yahshibu/nested-ner-tacl2020-transformers}}.

\subsection{Evaluation} We report labeled recall/precision/F1 scores based on \texttt{EVALB} \footnote{\url{https://nlp.cs.nyu.edu/evalb}} for constituency parsing;  span-level labeled recall/precision/F1 scores for nested NER. All reported results are averaged over three runs with different random seeds.

\subsection{Implementation details}
We use "bert-large-cased" \cite{devlin-etal-2019-bert} for PTB, ACE2004 and ACE2005; "bert-chinese-based" for CTB; and "biobert-large-cased-v1.1" \cite{Biobert} for GENIA. We use no other external resources (e.g., predicted/gold POS tags, external static word embedding). The hidden size of LSTM is set to 1000 for both the encoder and the decoder. 
We add dropouts in LSTM/MLP layers. The dropout rate is set to 0.33. The hidden and output sizes of all MLPs are set to 500.  The value of gradient clipping is set to 5. 
% We remove sentences of length $> 200$ to accelerate training. 
The number of training epochs is set to 10 for PTB, CTB, GENIA; 50 for ACE2004/2005.
% because they have fewer training sentences and thus need more epochs to converge. 
We use Adam \cite{DBLP:journals/corr/KingmaB14} as the optimizer with $\beta_1 = 0.9$ and $\beta_2 =0.9$. The maximal learning rate is set to $5e-5$ for BERT and $2.5e-3$ for all other components. 
We use the first $10\%$ epochs to linearly warmup the learning rates of each components to their maximum value and gradually decay them to zero for the rest of epochs. We batch sentences of similar lengths to make full use of GPUs and the number of tokens in a single batch is set to 3000.

\begin{table}[tb!]
    \centering 
    \scalebox{0.9}{
    \begin{tabular}{lccc}
        \toprule 
        {\bf Model} &   {\bf P} & {\bf R} & {\bf F}  \\
        \midrule
        \citet{kitaev-etal-2019-multilingual} \texttt{[S]} & 95.46 & 95.73 & 95.59 \\
        \citet{zhou-zhao-2019-head} \texttt{[S]} & 95.70 & 95.98 & 95.84\\
        \citet{TreeCRF} \texttt{[S]} & 95.85 & 95.53 & 95.69 \\
        \citet{IncrementalParser} \texttt{[T]} & 96.04 & 95.55 & 95.79\\ 
        \citet{nguyen-etal-2020-efficient} \texttt{[S]} & - & - & 95.48 \\ 
        \citet{wei-etal-2020-span} \texttt{[S]} & 95.5 & 96.1 & 95.8 \\
        \citet{tian-etal-2020-improving} \texttt{[S]} & 96.09 & 95.62 & 95.86 \\
        \citet{xin-etal-2021-n} \texttt{[S]}& \textbf{96.29} & 95.55 & 95.92 \\
        \citet{nguyen-etal-2021-conditional} \texttt{[Q]} & - & - & 95.7 \\ 
        \citet{DBLP:journals/corr/abs-2109-12814} \texttt{[S]} & 95.70 & \textbf{96.14} & 95.92 \\  
        \midrule
        Ours \texttt{[Q]} & 96.19 & 95.83 & \textbf{96.01}\\
       \bottomrule 
    \end{tabular}}
    \caption{Results on PTB. All models use BERT as encoders. \texttt{S}: span-based methods. \texttt{T}: transition-based methods. \texttt{Q}: seq2seq-based methods. P: labeled precision. R: labeled recall. F: labeled F1. }
    \label{tab:ptb}
\end{table}

\begin{table}[tb!]
    \centering 
    \begin{tabular}{lccc}
        \toprule 
        {\bf Model}   & {\bf P} & {\bf R} & {\bf F}  \\
        \midrule
        \citet{TreeCRF} \texttt{[S]} & \textbf{91.73} & \textbf{91.38} & \textbf{91.55}\\
        \midrule
        Ours \texttt{[Q]} &  91.66 & 91.31 & 91.49 \\
       \bottomrule 
    \end{tabular}
    \caption{Results on CTB7. All models use BERT as encoders. }
    \label{tab:ctb}
\end{table}

\section{Main result}
On both PTB and CTB, we find incorporating $E_{y_{t-1}}$ in Eq. \ref{eq:1} leads to a slightly inferior performance (-0.02 F1 score on PTB and -0.05 F1 score on CTB), so we report results without this input feature. 

Table \ref{tab:ptb} shows the results on PTB test set. Our method achieves 96.01 F1 score, outperforming the method of \citet{nguyen-etal-2021-conditional} by 0.31 F1 and having the same worst-case $O(n^2)$ parsing time complexity as theirs \footnote{In their paper, they claim an $O(n)$ time complexity, which treats the complexity of a single pointing operation as O(1). This calculation, however, assumes full GPU parallelization. Without parallelization, their method has a worst-case $O(n^2)$ time complexity as ours.}. It also outperforms all span-based methods, obtaining the state-of-the-art performance among all BERT-based models while enjoying a lower parsing complexity. 

Table \ref{tab:ctb} shows the results on CTB7. Our method obtains 91.49 F1 score, which is comparable to the method of \citet{TreeCRF} but has a lower complexity (worst-case $O(n^2)$ vs. $O(n^3)$). 

Table \ref{tab:ner} shows the results on three benchmark dataset on nested NER. We find that incorporating $E_{y_{t-1}}$ is important, leading to +0.67 F1 score and +0.52 F1 sore on ACE2004 and ACE2005, respectively. 
Although our method underperforms two recent state-of-the-art methods: \citet{shen-etal-2021-locate} and \citet{seq2set}, we find it has a competitive performance to other recent works \cite{wang-etal-2021-nested, yan-etal-2021-unified-generative, TreeCRFNER}. The most comparable one is the method of \citet{TreeCRFNER}, which belongs to parsing-based methods as ours. They adapt a span-based constituency parser to tackle nested NER using the CYK algorithm for training and inference.
Our model outperforms theirs by 0.34 F1 and 0.13 F1 scores on ACE2004 and ACE2005 and has a similar performance to theirs on GENIA, meanwhile enjoying a lower inference complexity.

\section{Analysis}

\begin{figure}[tb!]
\resizebox{\linewidth}{!}{%
\begin{tikzpicture}
  \begin{axis}[
    %   x tick label style={
    %     /pgf/number format/1000 sep=},
      ylabel={\small{F1 score}},
      legend style={
        font=\scriptsize,
        cells={anchor=west}
      },
      legend pos=north west,
      symbolic x coords={1-10,11-20,21-30,31-40,>40},
      xtick=data,
      ybar,
    ]
    \addplot coordinates {(1-10,96.24) (11-20, 95.00) (21-30, 95.91) (31-40, 97.07) (>40, 96.85)};
    \addplot coordinates {(1-10,96.10) (11-20, 94.80) (21-30, 95.60) (31-40, 96.53) (>40, 96.33)};
  \legend{Ours, \citet{nguyen-etal-2021-conditional}}
  \end{axis}
 \end{tikzpicture}
 }%

    \caption{F1 scores against constituent span length on PTB test set.}
        \label{fig:span_length}
\end{figure}
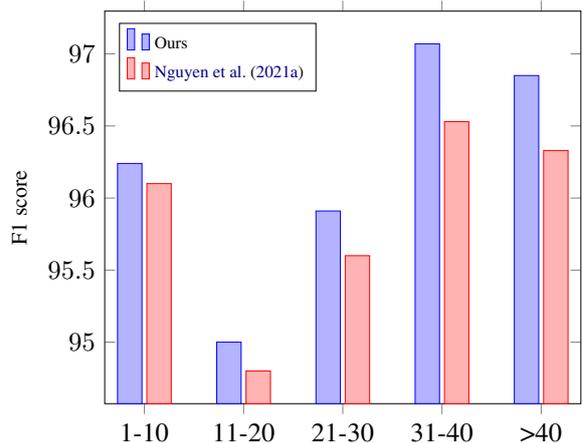

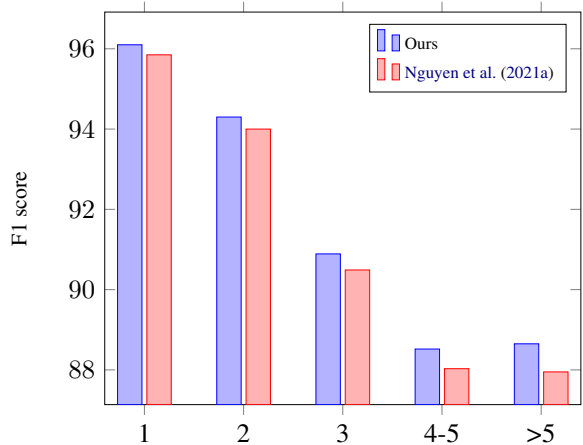
\begin{figure}[tb!]
\resizebox{\linewidth}{!}{%
\begin{tikzpicture}
  \begin{axis}[
    %   x tick label style={
    %     /pgf/number format/1000 sep=},
      ylabel={\small{F1 score}},
      legend style={
        font=\scriptsize,
        cells={anchor=west},
      },
      legend pos=north east,
      symbolic x coords={1,2,3,4-5,>5},
      xtick=data,
      ybar,
    ]
    \addplot coordinates {(1,96.10) (2, 94.30) (3, 90.89) (4-5, 88.52) (>5, 88.65)};
    \addplot coordinates {(1,95.85) (2, 94.00) (3, 90.49) (4-5, 88.03) (>5, 87.95)};
  \legend{Ours, \citet{nguyen-etal-2021-conditional}}
  \end{axis}
 \end{tikzpicture}
 }%

    \caption{F1 scores on constituent nodes with different numbers of children on PTB test set.}
        \label{fig:child}
\end{figure}

\begin{table*}[tb!]
    \centering 
    % \small
    \begin{tabular}{lccccccccc}
        \toprule 
    {\bf Model}    & \multicolumn{3}{c}{{\bf ACE2004}} & \multicolumn{3}{c}{{\bf ACE2005}} & \multicolumn{3}{c}{{\bf GENIA}}\\
        & P & R & F & P & R & F & P & R & F \\
        \midrule 
    \citet{shibuya-hovy-2020-nested} & 84.71 & 83.96 & 84.33 & 82.58 & 84.29 & 83.42 & 79.92 & 76.55 & 78.20 \\
    \citet{wang-etal-2020-pyramid} & 86.08 & 86.48 & 86.26 & 83.95 & 85.39 & 84.66 & 79.45 & 78.94 & 79.19 \\
    \citet{wang-etal-2021-nested} & 86.27 & 85.09 & 85.68 & 85.28 & 84.15 & 84.71 & 79.20 & 78.16 & 78.67 \\ 
    \citet{TreeCRFNER} & 86.7 & 86.5 & 86.6 & 84.5 & 86.4 & 85.4 & 78.2 & 78.2 & 78.2\\ 
    \citet{DBLP:conf/aaai/XuHF021} & 86.9 & 85.8 & 86.3 & 85.7 & 85.2 & 85.4 & 80.3 & 78.9 & 79.6 \\ 
    \citet{yan-etal-2021-unified-generative} & 87.27 & 86.41 & 86.84 & 83.16 & 86.38 & 84.74 & 78.57 & 79.3 & 78.93 \\ 
    \citet{shen-etal-2021-locate} & 87.44 & 87.38 & 87.41 & 86.09 & \textbf{87.27} & 86.67 & \textbf{80.19} & \textbf{80.89} & \textbf{80.54}\\ 
    \citet{seq2set} & 88.46 & 86.10 & 87.26 & \textbf{87.48} & 86.64 & \textbf{87.05} & 82.31 & 78.66 & 80.44\\ 
    \midrule 
    Ours  &  86.60 & 87.28 & 86.94   & 84.61 & 86.43 & 85.53 & 78.08 & 78.26 & 78.16\\ 
    \quad w.o. $E_{y_{t-1}}$ in Eq.\ref{eq:1} & 85.66 & 86.88 & 86.27 & 83.75 & 86.31&85.01 & 78.46&77.97&78.22\\
     \bottomrule 
    \end{tabular}
    \caption{Results on ACE2004, ACE2005 and GENIA. All models use BERT as encoders.}
    \label{tab:ner}
\end{table*}

\paragraph{Error analysis.}  As we discussed previously, bottom-up parsing can make use of the subtree features when predicting parent spans, so it is expected to have higher F1 scores on longer spans. To verify this, we plot Fig. \ref{fig:span_length} to show the changes of F1 scores with different constituent span lengths on the PTB test set. We can see that our method consistently outperforms the method of \cite{nguyen-etal-2021-conditional} on all span lengths, but our advantage is most prominent for spans of length >30,  which verifies our conjecture.   In Fig. \ref{fig:child}, we can see that when a constituent has multiple children (>3), our method performs much better than that of \cite{nguyen-etal-2021-conditional}, which validates the benefit of \textit{n-ary tree parsing}. An intuitive explanation of this benefit is that our method predicts n-ary branching structures in a single step, whereas theirs needs multiple steps, which is more error-prone.

\paragraph{Effect of beam search.} We also tried beam search but observed very slight improvement or even worse performance (e.g., +0.05 F1 score on PTB and -0.03 F1 score on CTB when we use a beam size 20). Hence we report all results using greedy decoding for simplicity. This suggests that greedy decoding can yield near-optimal solutions, indicating that our model is less prone to the error propagation problem. 

\paragraph{Effect of training loss.} As discussed in Sec. 2.3, we find that explicitly considering the structural consistency constraints when normalizing is harmful (-0.12 F1 score on PTB, -0.10 F1 score on CTB).  We speculate that not enforcing the constraints during training can help the model to learn the constraints implicitly, which is helpful for the model to generalize better on the unseen test set. Notably, \citet{nguyen-etal-2021-conditional} also adopt this strategy, i.e., normalizing over all boundaries.

\paragraph{Speed.} Similar to \citet{nguyen-etal-2021-conditional}, the training process (i.e., teacher forcing)
can be fully parallelized without resorting to structured inference, which could be compute-intensive or hard to parallelize.
On PTB, it takes only 4.5 hours to train the model using BERT as the encoder with a single Titan V GPU. As for parsing,  our method has the same parsing complexity as \citet{nguyen-etal-2021-conditional}, i.e., worst-case $O(n^2)$.
Table \ref{table:speed} shows the speed comparison on parsing the PTB test set (we report values based on a single Titan V GPU and not using BERT as encoder following \citet{nguyen-etal-2021-conditional}). We report the average number of pointing actions in Appendix A.2.
\begin{table}[tb]
\centering
\scalebox{0.7}{
\begin{tabular}{lcc}  
%\toprule
\textbf{System} & \bf{Speed (Sents/s)} & \bf{Speedup}\\
\midrule
\citet{petrov-klein-2007-improved} (Berkeley)          & 6   &1.0x\\
\citet{zhu-etal-2013-fast}(ZPar)                                      & 90  &15.0x\\
\citet{stern-etal-2017-minimal}                                   & 76  &12.7x\\
\citet{shen-etal-2018-straight}                                   & 111 &18.5x\\
\citet{nguyen-etal-2020-efficient}                                  & 130 &21.7x \\
\citet{zhou-zhao-2019-head}                                          & 159 &26.5x\\
\citet{wei-etal-2020-span}                                          & 220 &36.7x\\
% \citet{kitaev-klein-2018-constituency}                     & 332 &55.3x\\
\citet{gomez-rodriguez-vilares-2018-constituent} & 780 &130x\\
% \citet{ijcai2020-560}                                                    & 1092 &182x\\
\citet{kitaev-klein-2018-constituency} (GPU)                    & 830 &138.3x\\
\citet{TreeCRF}                                                    & 924 &154x\\
\citet{nguyen-etal-2021-conditional}                                             & 1127 &187.3x\\
Ours &  855 &  142.5x \\
\midrule
\bottomrule
\end{tabular}
}
% \vspace{-0.5em}
\caption{Speed comparison.}
\label{table:speed}
\end{table}

% Effect of decoder input features}
%  We show the effect of the use of $E_{y_{t-1}}$ at Table. We can see that the use of $E_{y_{t-1}}$ is crucial in nested NER.

% \begin{table}[t]
%     \centering 
%     \small
%     \begin{tabular}{lcccccc}
%         \toprule 
%         & {\bf PTB} & {\bf CTB}\\
%         \midrule 
%       \textbf{w/} $E_{y_{t-1}}$ & 97.24 & 95.73\\
%       \textbf{w/o} $E_{y_{t-1}}$ & 93.28 & 92.20\\
%       \bottomrule 
%     \end{tabular}
%     \caption{The influence of training loss function on PTB and CTB.}
%     \label{abl1}
% \end{table}

% \begin{table}[t]
%     \centering 
%     \small
%     \begin{tabular}{lccccccc}
%         \toprule 
%         & {\bf ACE2004} & {\bf ACE2005} & {\bf GENIA}\\
%         \midrule 
%       \textbf{w/} $E_{y_{t-1}}$ & 97.24 & 95.73\\
%       \textbf{w/o} $E_{y_{t-1}}$ & 93.28 & 92.20\\
%       \bottomrule 
%     \end{tabular}
%     \caption{The influence of training loss function on PTB and CTB.}
%     \label{abl1}
% \end{table}

\section{Related Work}
\paragraph{Constituency parsing.} There are many methods to tackle constituency parsing, such as transition-based methods \cite{dyer-etal-2016-recurrent,cross-huang-2016-span, liu-zhang-2017-order, IncrementalParser}, span-based methods \cite{stern-etal-2017-minimal, kitaev-klein-2018-constituency, kitaev-etal-2019-multilingual, TreeCRF, wei-etal-2020-span, nguyen-etal-2020-efficient, xin-etal-2021-n}, sequence-to-sequence (seq2seq)-based methods \cite{DBLP:conf/nips/VinyalsKKPSH15, fernandez-gonzalez-gomez-rodriguez-2020-enriched}, sequence-labeling-based methods \cite{gomez-rodriguez-vilares-2018-constituent, vilares-etal-2019-better, kitaev-klein-2020-tetra}, among others. 

Our work belongs to the category of seq2seq-based methods. Previous seq2seq models linearize constituency trees into  bracket sequences \cite{DBLP:conf/nips/VinyalsKKPSH15} or shift-reduce action sequences \cite{DBLP:conf/aaai/MaLTZS17, fernandez-gonzalez-gomez-rodriguez-2020-enriched}. However, they may produce invalid outputs and their performance lags behind span-based methods.  Recently, seq2seq models linearize constituency trees into sequences of spans in pre-order  \cite{nguyen-etal-2021-conditional} or in in-order \cite{InorderChart}. Our method generates sequences of spans in post-order instead, which has the advantage of utilizing  rich subtree features and performing direct n-ary tree parsing.

% Our work is also closely related to previous transition-based bottom-up parsers. Compared with them, the use of pointer networks enables our method to capture information from the whole input sentence. Moreover, our method makes global pointing decisions to directly generate a constituent of any length in one step, while transition-based methods make smaller local decisions, thus needing multiple steps to generate a long constituent, which is more error-prone.

Binarization is \textit{de facto} in constituency parsing, but there is a recent trend toward n-ary parsing.  
 Previous span-based methods adopt either explicit binarization \cite{TreeCRF} or implicit binarization \cite{stern-etal-2017-minimal, kitaev-klein-2018-constituency}. Although the implicit binarization strategy eliminates the need for binarization in training, it can only output binary trees during decoding. 
\citet{xin-etal-2021-n} propose an n-ary-aware span-based method by defining semi-Markov processes on each parent span so that the transition scores of adjacent sibling child-spans are explicitly considered in parsing. \citet{DBLP:journals/ai/Fernandez-Gonzalez19, IncrementalParser} propose novel transition systems to model n-ary trees. Our method outputs n-ary trees without the need for binarization via a novel pointing mechanism.

\paragraph{Parsing with pointer networks.} Pointer Networks \cite{PointerNet} are introduced to the parsing literature by \citet{ma-etal-2018-stack} and quickly become popular in various parsing subtasks because they are flexible to predict various trees/graphs and can achieve very competitive performance. \citet{ma-etal-2018-stack}  linearize a dependency tree in a top-down depth-first and inside-out manner and use a pointer network to predict the linearized dependency tree, which is then extended by  \citet{lin-etal-2019-unified} to discourse parsing. \citet{liu-etal-2019-hierarchical} add shortcuts between the decoder states of the previously generated parents/siblings to the current decoder states in both dependency and discourse parsing. \citet{fernandez-gonzalez-gomez-rodriguez-2019-left} propose a left-to-right dependency parser that predicts the heads of each word autoregressively, and later, they propose right-to-left and outside-in variants \cite{BottomUpPtr}.  They also adapt the left-to-right dependency parser to semantic dependency parsing (which predicts acyclic graphs instead of trees) \cite{fernandez-gonzalez-gomez-rodriguez-2020-transition}, discontinuous constituency parsing (by treating discontinuous constituency trees as augmented dependency trees) \cite{DBLP:conf/aaai/Fernandez-Gonzalez20}, and joint dependency and constituency parsing \cite{DBLP:journals/corr/abs-2009-09730}. They use a pointer network to reorder the sentence to reduce discontinuous constituency parsing to continuous constituency parsing \cite{DBLP:journals/corr/abs-2104-06239}.
\citet{ nguyen-etal-2021-conditional, nguyen-etal-2021-rst} cast (discourse) constituency/RST  parsing as conditional splitting and use pointer networks to select the splitting points. \citet{zhou-etal-2021-amr} propose an action-pointer network for AMR parsing.
% Beyond parsing, there are many works using pointer networks in other NLP structured prediction tasks \cite{DBLP:conf/aaai/NayakN20, chen-etal-2021-jointly, mukherjee-etal-2021-paste,DBLP:conf/aaai/0001JLLRL21,DBLP:conf/aaai/0001LLJ21}.

\paragraph{Nested NER.}  There are also many methods to tackle nested NER, such as hypergraph-based methods \cite{lu-roth-2015-joint, katiyar-cardie-2018-nested, wang-lu-2018-neural}, sequence-labeling-based methods \cite{shibuya-hovy-2020-nested,wang-etal-2021-nested}, parsing-based methods \cite{finkel-manning-2009-nested, wang-etal-2018-neural-transition, TreeCRFNER}, layered methods \cite{fisher-vlachos-2019-merge, wang-etal-2020-pyramid, luo-zhao-2020-bipartite}, span-based methods \cite{yu-etal-2020-named, li-etal-2021-span}, object-detection-based methods \cite{shen-etal-2021-locate,seq2set} etc. 

Our work belongs to the category of parsing-based methods.  \citet{finkel-manning-2009-nested} insert named entities into a constituency tree and use a discriminative parser \cite{finkel-etal-2008-efficient} for learning and prediction. 
\citet{wang-etal-2018-neural-transition} adapt a shift-reduce transition-based parser to output a constituency forest instead of a constituency tree for nested NER. 
% \citet{yu-etal-2020-named} adopt ideas from the biaffine dependency parser \cite{Biaffine} to handle both nested and flat named entities.
\citet{TreeCRFNER} adapt a span-based neural TreeCRF parser, treat nested named entities as the observed parts of a partially-observed constituency tree and develop a masked inside algorithm to marginalize all unobserved parts for maximizing the probability of the observed named entities. Our method has a better performance as well as a lower time complexity than \citet{TreeCRFNER}. Recently, \citet{Lou2022Nested} extend the work of \citet{TreeCRFNER}, casting nested NER 
to lexicalized constituency parsing for leveraging headword information. They achieve a higher performance at the cost of a higher parsing complexity, i.e., $O(n^4)$. 

\section{Discussion and future work}
In the deep learning era, global optimization on trees becomes less important in both training and decoding. \citet{teng-zhang-2018-two} show that a span-based model trained with a local span classification loss performs well in conjunction with CYK decoding. \citet{wei-etal-2020-span, nguyen-etal-2020-efficient} show that top-down greedy decoding performs comparably. In this work we have shown that greedy decoding works well. Thus it would also be a fruitful direction to design more powerful neural decoders which can leverage more subtree information and can maintain structural consistency. Also, it is a fruitful direction to devise more powerful span representations.

\section{Conclusion}
In this work we have presented a novel pointing mechanism and model for bottom-up constituency parsing, which allows n-ary tree parsing in linear steps. Experiments on multiple datasets show the effectiveness of our methods in both constituency parsing and nested NER.
\section*{Acknowledgments}
We thank the anonymous reviewers for their constructive comments. This work was supported by the National Natural Science Foundation of China (61976139).

\bibliography{anthology,custom}
\bibliographystyle{acl_natbib}

\appendix

\section{Appendix}
\subsection{Uniqueness of oracle}
If there are two oracles $o_1$ and $o_2$ outputting the same tree. Their parsing configuration is $(c_1, A_1, p_1, S_1)$ and $(C_2, A_2, p_2, S_2)$, respectively.  Assume that the first $k$th pointing actions of $o_1$ and $o_2$ are the same (so they share the same cursor $c$) and the $k+1$th action is $(c \rightarrow a_1, y_1)$ and $(c \rightarrow a_2, y_2)$ respectively. We enumerate all possibilities:
\begin{itemize}
\item $a_2 < c < a_1$, then $(a_2, c)$  exists in $S_2$. $c_1$ would be updated to  $a_1$, so thereafter the endpoint of the generated span is $\ge a_1$, thus $(a_2, c)$ cannot exist in $S_1$ since $c < a_1$.
\item $a_1 < c < a_2$, then $(a_1, c)$  exists in $S_2$. Similar to the previous case, we can conclude that $(a_1, c)$ cannot exist in $S_1$.
\item $a_1 < a_2 < c$, then $(a_2, c)$ exists in $S_2$, but $a_2 \not\in A_1$ for all remaining steps, thus $(a_2, c)$ cannot exist in $S_1$. 
\item $a_2 < a_1 < c$. This is similar to the previous case.
\item $c < a_1 < a_2$, then $(a_1, c)$ exists in $S_1$, but $a_1 \not\in A_2$ for the remaining steps, thus $(a_1, c)$ cannot exist in $S_2$.
\item $c < a_2 < a_1$.  This is similar to the previous case.
\end{itemize}
Hence there is exact one oracle and we have proved it by contradiction.

\begin{figure}[tb!]
    \centering
    \includegraphics[width=1\linewidth]{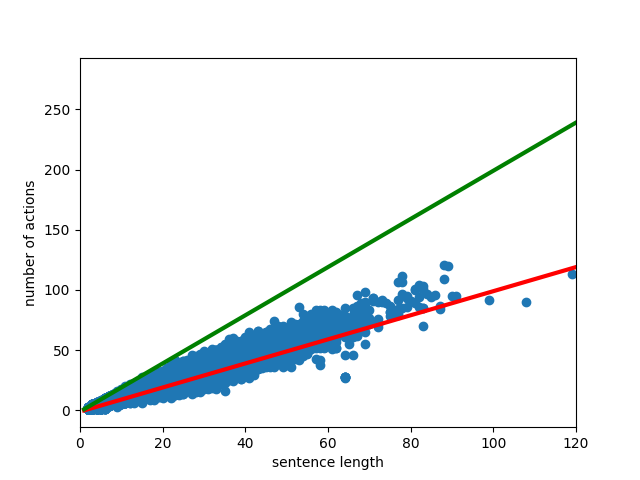}
    \caption{The number of actions with different sentence lengths in  PTB}
    \label{fig:ptb_count}
\end{figure}

\begin{figure}[tb!]
    \centering
    \includegraphics[width=1\linewidth]{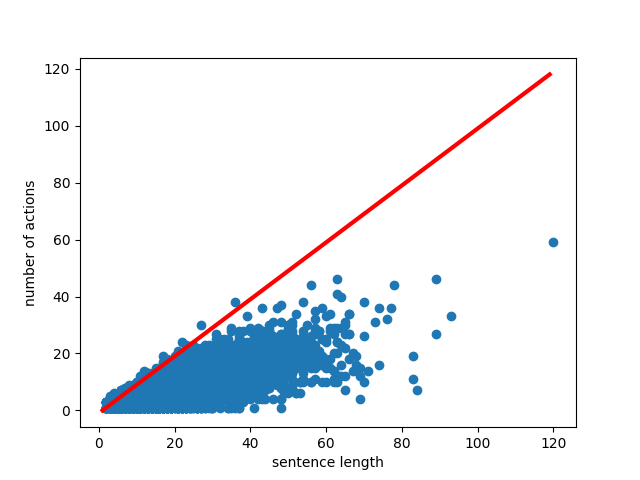}
    \caption{The number of actions with different sentence lengths in  ACE2004}
    \label{fig:ace_count}
\end{figure}

\subsection{Number of actions}

The system of \citet{nguyen-etal-2021-conditional} needs exact $n-1$ actions to parse a length-$n$ sentence. While our model requires $2n-1$ actions in the worst case because we generate one span at each step and there are at most $2n-1$ spans if the corresponding constituency tree is a full binary tree. So there is a concern that our model needs twice time to parse. Empirically, since the constituency trees in the treebank are not full binary trees in most cases, we need less than $2n-1$ steps to parse. Fig. \ref{fig:ptb_count} shows the number of actions needed to parse with different sentence lengths in PTB training set. The red line is $y=x-1$ and the green line is $y=2x-1$. In average, our method needs 1.13 actions per token, \citet{nguyen-etal-2021-conditional} needs 0.96 action per token. So, our method is around 20\% slower than theirs. Fig. \ref{fig:ace_count} shows the case in nested NER. We only need 0.40 action per token since the spans in nested NER is more \textit{sparse} than that in constituency parsing.  Our method is expectedly faster than other parsing-based methods in nested NER, such as the transition system of \citet{wang-lu-2018-neural}, which needs at least one action per word; and the span-based method of \citet{TreeCRFNER}, which needs cubic time for CYK parsing.

\label{sec:appendix}

\end{document}